\documentclass[11pt, a4paper, onecolumn, thu]{thuc3i}

\usepackage[authoryear, sort&compress, square]{natbib}
\usepackage{fontawesome5}
\usepackage[most]{tcolorbox}
\definecolor{AbstractBgColor}{HTML}{F4F7FB}

\usepackage{mathtools}
\usepackage[capitalise,noabbrev]{cleveref}

\graphicspath{{figures/}}

\newcommand{\sysname}{MARCH}

\setheadertext{MARCH}
\setheaderlogos{%
  \includegraphics[height=2em]{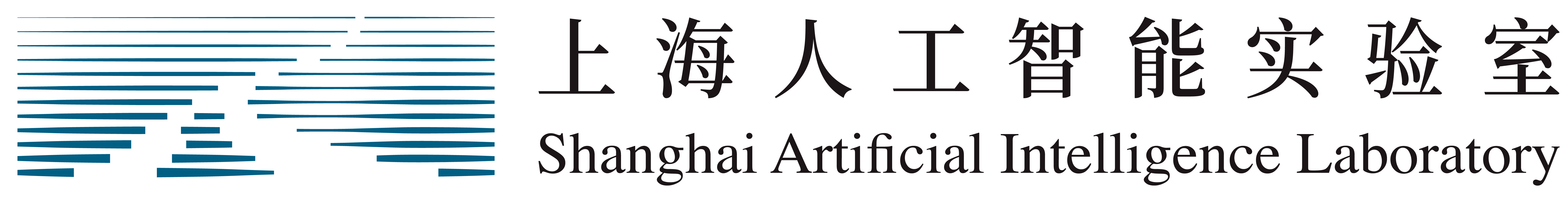}%
  \hspace{12pt}%
\includegraphics[height=22pt]{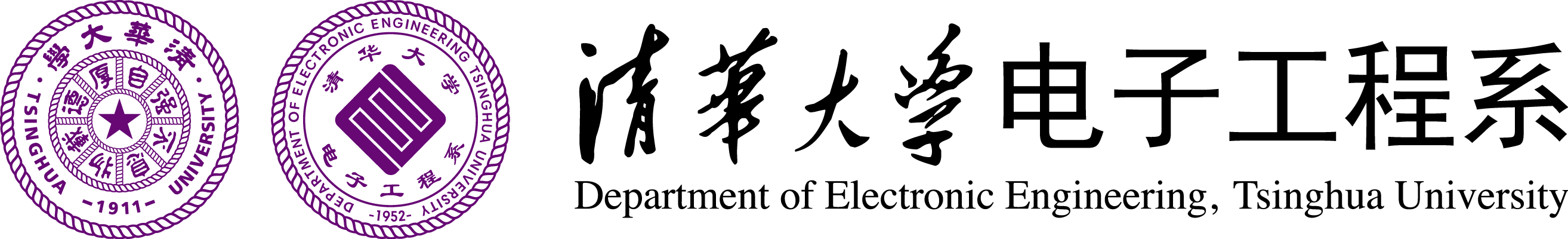}%
}

\title{MARCH: Scaling Recurrent Memory with Content-Routed State Anchors}
\author{Ming Zhang$^{1,2*}$, Kaisen Yang$^{2*}$, Shu Yu$^{1,3}$, Ermo Hua$^{1,2}$, Ning Ding$^{1,2}$, Xia Hu$^{1}$, Bowen Zhou$^{1,2}$, \qquad Chaochao Lu$^{1\dagger \sharp}$, Youbang Sun$^{1,2\ddagger\sharp}$
    \vspace{1mm} \\
    $^1$ Shanghai AI Laboratory \quad
    $^2$ Tsinghua University \quad
    $^3$ Fudan University \\
    \textbf{$^*$ Equal Contribution.}~~  
    \textbf{$^\dagger$ Project Lead.}~~ 
    \textbf{$^\ddagger$ Technical Lead.}~~
    \textbf{$^\sharp$ Corresponding Authors.}
    \vspace{1mm} \\
    
}
\reportnumber{}

\begin{abstract}
  Transformers owe much of their strong long-context retrieval capability to a token-level memory that grows with context length. 
This flexibility, however, incurs a quadratic computation complexity during training and a key--value cache that grows linearly during autoregressive inference. 
Recurrent alternatives offer efficient decoding by compressing the entire history into a fixed-size state, but often underperform on recall-intensive tasks since earlier associations usually get overwritten by subsequent updates, and only the most recent contextual information is retained. 
In this paper, we introduce \textbf{M}emory-\textbf{A}nchor \textbf{R}outing across \textbf{C}ontext \textbf{H}istory (\textbf{MARCH}), a network architecture that effectively scales state-space models beyond a fixed-size dimension, while maintaining computational efficiency over long-sequences.
MARCH periodically caches cumulative recurrent-state checkpoints as \emph{state anchors} and associates each anchor with a compact, content-conditioned anchor key. 
This lets MARCH maintain a memory bank, which can grow as context length increases, providing a controllable trade-off between historical resolution and memory cost. At each token, MARCH produces an anchor query to attend all causally available state anchors, and the output is calculated as an attention-style aggregation over all historical anchors along the current state.
We show that after standard pretraining, MARCH consistently outperforms multiple linear attention variants across commonsense reasoning, LongBench, and in-context retrieval.
These results demonstrate that content-routed state caching substantially strengthens recurrent long-range memory while preserving its native computation path.

\end{abstract}

\begin{document}
\begin{tcolorbox}[
    colback=AbstractBgColor, 
    colframe=AbstractBgColor, 
    arc=5pt,                  
    auto outer arc,
    boxrule=0pt,              
    left=8pt, right=8pt, 
    top=4pt, bottom=4pt,      
    parbox=false,
    width=\textwidth,
    before skip=-800pt,        
    after skip=0pt, 
    enlarge top by=-12pt
]
\vspace{0.5em}

\maketitle

\end{tcolorbox}

\vspace{0.5em}

\begin{figure}[!ht]
  \centering
  \includegraphics[width=\textwidth]{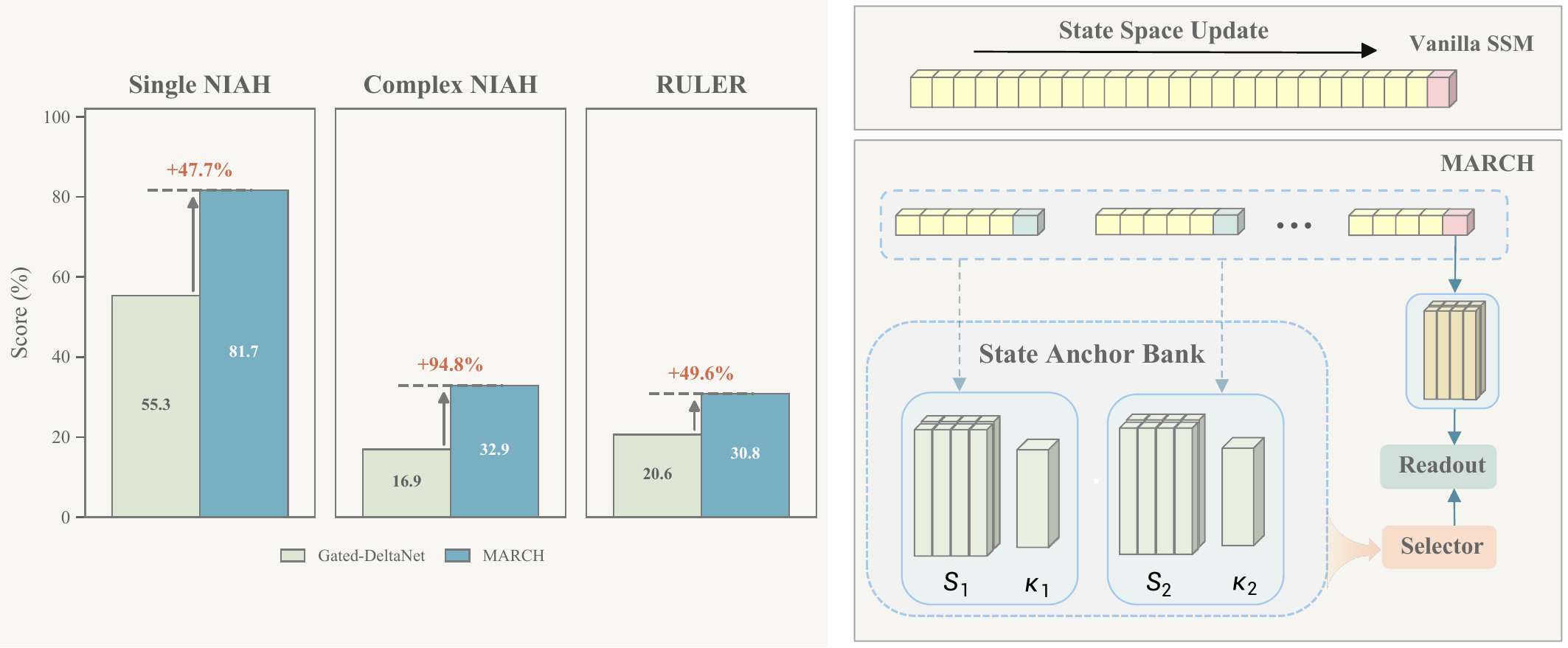}
  \caption{Overview of \sysname{}. Left: long-context retrieval performance at
  a context length of 8K for \sysname{} and Gated DeltaNet. Right: State anchors enable content-based retrieval of historical recurrent states.}
  \label{fig:march-overview}
\end{figure}

\newpage
\section{Introduction}
\label{sec:introduction}
Large language models (LLMs) have demonstrated remarkable capabilities across a wide range of language understanding and generation tasks. However, many real-world applications—including long-document understanding, multi-turn interaction, and in-context learning—require models to integrate information distributed across extended sequences~\citep{bai2025longbench,kwan2024mt,zou2025many}. Supporting such contexts involves more than simply increasing the number of input tokens: models must retain relevant information across long intervening spans and reliably retrieve it when needed~\citep{liu2023lost,hsieh2024ruler}. Effective long-context modeling therefore hinges on a model's ability to manage memory—determining what information to preserve, how to represent it, and when to retrieve it~\citep{wang2023augmenting,behrouz2024titans}.  A useful
perspective is to view a sequence model as a memory system with two basic
operations:
\emph{writing}, which incorporates each new input into memory, and
\emph{reading}, which retrieves information relevant to the current input
\citep{behrouz2024titans}.  Under this view, standard
self-attention maintains a growing token-level memory in its key--value cache
\citep{vaswani2017attention}: it writes by appending each new key--value pair
without compressing the existing cache, and reads by matching the current
query against all stored keys and combining their associated values.  This
uncompressed, token-level memory provides a direct path to every
preceding token, enabling accurate recall of fine-grained details and distant
dependencies.  Its flexibility, however, entails quadratic computation during
training and a key--value cache that grows linearly with sequence length
during autoregressive inference, making self-attention increasingly costly as
context windows expand.

Linear attention and modern recurrent sequence models make the opposite
trade-off \citep{katharopoulos2020transformers,dao2024transformers}.
They compress the causal prefix into a fixed-size, matrix-valued recurrent state, update this state with each new input, and read from it by applying the current query.  This design enables constant-memory recurrent decoding, but the same compressed write that
makes it efficient also limits its long-range memory.  At each step,
information from the new token is written into a state already shared by the
entire history. Much recent work has therefore
focused on improving the write operation.  Selective state-space models such
as Mamba introduce input-dependent state transitions and forgetting
\citep{gu2023mamba}, whereas DeltaNet and Gated DeltaNet
use data-dependent delta-rule updates to revise existing associations before incorporating new information
\citep{schlag2021lineartransformersselfretrievalcompressive,yang2025gateddelta}. These mechanisms improve state
tracking and mitigate indiscriminate accumulation, but they do not eliminate
the underlying fixed-state bottleneck: the entire history must still share
one evolving state, and only its latest version remains available for
reading.  Indeed, although recent linear recurrent models can match or
surpass softmax attention in short-context settings, their performance often
degrades as the evaluation context grows
\citep{arora2024zoology,wang2026dynamic}.  Once an earlier association has
been weakened by forgetting or modified by subsequent writes, the model has no direct path to its earlier representation and cannot recover it from the latest state alone.

Recent works have relaxed the fixed-state bottleneck along two broad directions. One approach increases the memory capacity available at each step through partitioned sparse states, large memories with sparse reads and writes, or routed mixtures of independent states
\citep{pan25SSE,cabannes26SDM,du2025mom}. A second line preserves a temporally structured collection of compressed states through logarithmic hierarchies, adaptive state construction and merging, or recurrent-state caching
\citep{guo2026loglinear,wang2026dynamic,behrouz26memory}.
Collectively, these approaches demonstrate that expanding memory capacity or temporal coverage can improve long-range recall. 
However, with the exception of certain instances in \citep{behrouz26memory}, all of the existing works still maintain a finite or upper-constrained state space dimension, though the dimensionality of which is increased. 
Moreover, as multiple states become available, the primary bottleneck shifts from memory construction to memory retrieval: the model must determine which state retains the information most relevant to the current token. 
In particular, how to construct context-dependent representations for historical checkpoints that facilitate effective and efficient query-dependent retrieval remains underexplored.

In this work, we introduce \textbf{M}emory-\textbf{A}nchor \textbf{R}outing across \textbf{C}ontext \textbf{H}istory (\textbf{MARCH}), a memory-augmented recurrent
architecture that enables selective retrieval from earlier versions of recurrent memory. Without modifying the underlying recurrence, MARCH periodically preserves cumulative states as \emph{state anchors}, giving later tokens access to earlier versions of the evolving memory. Each anchor is associated with a compact learned descriptor, allowing the model to route each token to relevant historical states when additional context is needed and combine their contents with the current-state readout. MARCH thereby complements efficient recurrent processing with selective access to preserved historical memory. The resulting mechanism remains causal and is trained end to end with the standard language-modeling objective.

Our main contributions are summarized as follows:
\begin{itemize}
  \item \textbf{Routable memory-state bank.}
  We introduce MARCH, which expands fixed-state recurrence into a growing bank of historical memory states which increases its capacity as context grows, alleviating the single-state memory bottleneck without modifying the underlying recurrence.

  \item \textbf{Content-conditioned historical retrieval.}
 MARCH brings attention-style content routing to recurrent memory by applying a standard softmax over compact keys for a temporally sparse set of state anchors rather than token-level key--value pairs. A learned null route allows the model to suppress the historical branch when the current recurrent state is sufficient, while residual fusion preserves the original recurrent path and supports end-to-end training.

  \item \textbf{Extensive empirical validation.}
  We demonstrate consistent improvements over strong recurrent
  baselines across commonsense reasoning, LongBench, in-context
  retrieval, and NIAH evaluations, including robust extrapolation beyond
  the training context length.
\end{itemize}

\section{Preliminaries}
\label{sec:preliminaries}

\paragraph{Full and Linear Attention as Memory.}
Let $\mathbf{x}_t\in\mathbb{R}^{d}$ denote the hidden representation at
position $t$. The corresponding query, key, and value vectors are obtained
through learned linear projections:
\begin{equation}
  \mathbf{q}_t=\mathbf{W}_q\mathbf{x}_t,\qquad
  \mathbf{k}_t=\mathbf{W}_k\mathbf{x}_t,\qquad
  \mathbf{v}_t=\mathbf{W}_v\mathbf{x}_t,
\end{equation}
where
$\mathbf{W}_q,\mathbf{W}_k\in\mathbb{R}^{d_k\times d}$ and
$\mathbf{W}_v\in\mathbb{R}^{d_v\times d}$, such that
$\mathbf{q}_t,\mathbf{k}_t\in\mathbb{R}^{d_k}$ and
$\mathbf{v}_t\in\mathbb{R}^{d_v}$. Following the memory-system perspective adopted in prior work
\citep{behrouz2024titans}, we view a causal sequence mixer as an online
memory system. Let $\mathcal{M}_t$ denote the memory state after processing
the first $t$ tokens, with $\mathcal{M}_0$ denoting its initial state.
At each position, the current key--value pair is first written into memory,
after which the updated memory is queried using the current query:
\begin{equation}
  \mathcal{M}_t
  =
  \operatorname{Write}
  \bigl(\mathcal{M}_{t-1};\mathbf{k}_t,\mathbf{v}_t\bigr),
  \qquad
  \mathbf{o}_t
  =
  \operatorname{Read}
  \bigl(\mathcal{M}_t;\mathbf{q}_t\bigr),
  \label{eq:prelim-memory}
\end{equation}
where $\mathbf{o}_t\in\mathbb{R}^{d_v}$ denotes the memory readout at
position $t$. For causal softmax attention, the memory explicitly retains all projected
key--value pairs observed up to position $t$:
\begin{equation}
  \mathcal{M}_t
  =
  \bigl(\mathbf{K}_{\leq t},\mathbf{V}_{\leq t}\bigr),
\end{equation}
where $\mathbf{K}_{\leq t}\in\mathbb{R}^{t\times d_k}$ and
$\mathbf{V}_{\leq t}\in\mathbb{R}^{t\times d_v}$ stack the keys and values
row-wise, respectively. Writing appends
$(\mathbf{k}_t,\mathbf{v}_t)$ to these matrices, whereas reading performs
content-based retrieval:
\begin{equation}
  \mathbf{o}_t
  =
  \mathbf{V}_{\leq t}^{\top}
  \operatorname{softmax}
  \left(
    \frac{\mathbf{K}_{\leq t}\mathbf{q}_t}{\sqrt{d_k}}
  \right).
  \label{eq:prelim-full-attention}
\end{equation}
This explicit storage keeps individual tokens directly retrievable and enables
fine-grained retrieval from the entire causal prefix. However, processing a
sequence of length $T$ requires $\mathcal{O}(T^2)$ query--key interactions,
while autoregressive decoding maintains a key--value cache of size
$\mathcal{O}\!\left(T(d_k+d_v)\right)$ per attention head
\citep{vaswani2017attention}.

Linear attention instantiates the memory state $\mathcal{M}_t$ as a
fixed-size matrix $\mathbf{S}_t\in\mathbb{R}^{d_v\times d_k}$. Its write and
read operations are given by
\begin{equation}
  \mathbf{S}_t
  =
  \mathbf{S}_{t-1}
  +
  \mathbf{v}_t\mathbf{k}_t^{\top},
  \qquad
  \mathbf{o}_t
  =
  \mathbf{S}_t\mathbf{q}_t.
  \label{eq:prelim-la}
\end{equation}
Each write therefore adds a rank-one key--value association to the shared matrix, while each read retrieves a query-dependent superposition of the stored values. This enables
constant-memory recurrent decoding, but introduces interference as the
compressed history grows.

\paragraph{Gated DeltaNet (GDN).}
To mitigate the interference caused by the additive write rule of linear
attention, GDN retains the same state-based read operation
but introduces input-dependent retention and a targeted delta-rule write
\citep{yang2025gateddelta}:
\begin{equation}
  \mathbf{S}_t
  =
  \textcolor{red}{\alpha_t}\mathbf{S}_{t-1}
  +
  \textcolor{blue}{\beta_t}
  \left(
    \mathbf{v}_t
    -
    \textcolor{red}{\alpha_t}
    \mathbf{S}_{t-1}\mathbf{k}_t
  \right)
  \mathbf{k}_t^{\top},
  \qquad
  \mathbf{o}_t
  =
  \mathbf{S}_t\mathbf{q}_t,
  \label{eq:prelim-gdn}
\end{equation}
where $\alpha_t\in(0,1)$ is an input-dependent retention gate, and
$\beta_t\in[0,1]$ modulates the strength of the targeted delta update.
Despite its more adaptive state dynamics, GDN still compresses the entire causal history into a single fixed-size recurrent state. Because all associations share this evolving state, information weakened or overwritten by subsequent updates has no direct retrieval path, limiting reliable long-context recall.

\paragraph{Scaling recurrent memory.}
Recent work has sought to relax the fixed-state bottleneck along two broad
directions. Capacity-expansion methods enlarge the current recurrent state,
whereas temporal-expansion methods retain multiple versions of the state along
its trajectory:
\begin{equation}
  \begin{aligned}
    \mathcal{M}_t^{\mathrm{cap}}
    &\coloneqq
    \widetilde{\mathbf{S}}_t
    =
    \bigl[
      \mathbf{S}_t^{(1)}
      \mid \cdots \mid
      \mathbf{S}_t^{(P)}
    \bigr]
    \in\mathbb{R}^{d_v\times D_{\mathrm{mem}}},
    \qquad
    D_{\mathrm{mem}}
    =
    \sum_{p=1}^{P} d_p,
    \\
    \mathcal{M}_t^{\mathrm{temp}}
    &\coloneqq
    \bigl(
      \overline{\mathbf{S}}_{t,1},
      \ldots,
      \overline{\mathbf{S}}_{t,M_t}
    \bigr),
    \qquad
    \overline{\mathbf{S}}_{t,m}
    \in\mathbb{R}^{d_v\times d_k},
    \qquad
    1
    \leq
    \tau_{t,1}
    <
    \cdots
    <
    \tau_{t,M_t}
    \leq t.
  \end{aligned}
  \label{eq:prelim-state-scaling}
\end{equation}
In the capacity formulation, $P$ is the fixed number of state partitions and
$D_{\mathrm{mem}}$ is their total memory dimension. Such methods increase
$D_{\mathrm{mem}}$ while using sparse access to keep computation tractable
\citep{pan25SSE,cabannes26SDM}. In the temporal formulation, $M_t$ is the
number of retained state representations, each associated with a temporal
boundary $\tau_{t,m}$. These methods preserve states from distinct temporal
regions or earlier stages of the recurrent trajectory
\citep{guo2026loglinear,wang2026dynamic,behrouz26memory}. MARCH follows the latter direction by retaining cumulative snapshots of a
continuously evolving recurrent state.
\section{Method}
\label{sec:method}

\begin{figure}[t]
  \centering
  \includegraphics[width=\textwidth]{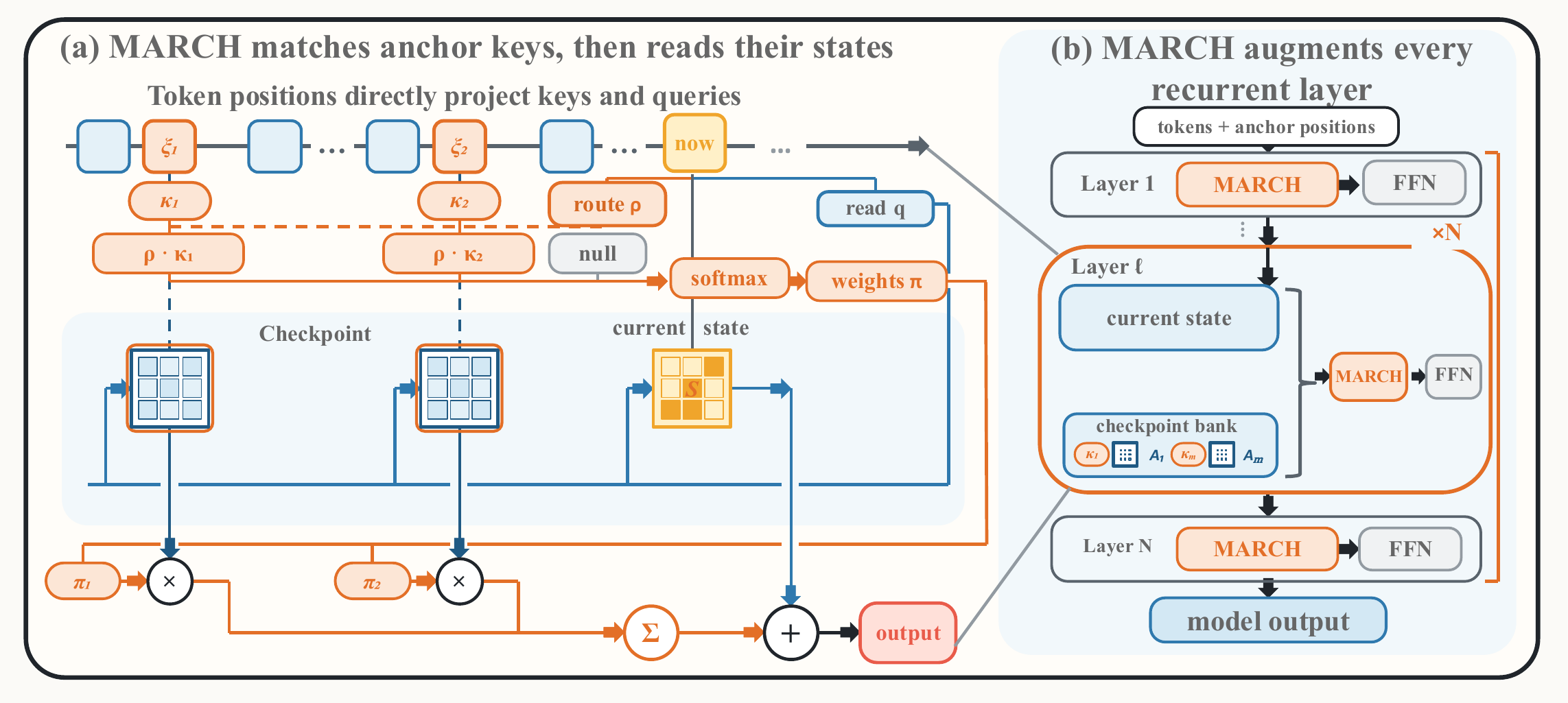}
  \caption{Architecture of \sysname{}. Left: Text tokens update one continuous
  Gated DeltaNet state, while periodic checkpoints form state anchors.
  Token-dependent routing combines causally visible anchors, and the resulting
  historical readout is added to the current-state readout. Right: \sysname{}
  augments every recurrent layer and rebuilds state-aware routing keys across
  layers.}
  \label{fig:march-architecture}
\end{figure}

Figure~\ref{fig:march-architecture} illustrates MARCH, a content-routed
recurrent memory framework that enables selective retrieval from earlier
versions of an evolving recurrent state. Rather than routing over predefined
state indices or temporal scales, MARCH matches each query against individual
historical states based on their contents. As tokens are processed, MARCH
periodically checkpoints the cumulative recurrent state, producing a bank of
state anchors. Each checkpoint is paired with an occurrence of a shared
learned anchor token, whose hidden representation yields a compact \emph{routing key}. For each text token, a \emph{routing query} scores all causally visible anchors
alongside a learned null option, allowing the model to use historical memory
only when useful. The resulting routing probabilities define a weighted
combination of the visible anchor states, which is read using the token's
standard recurrent query. This historical readout is added to the
current-state readout, preserving the native recurrent path while introducing
a content-dependent route to earlier memory. Together, state anchoring and
content-routed retrieval turn the otherwise transient state trajectory into a
persistent source of long-range memory.

\subsection{Continuous Recurrent-State Anchoring}
\label{sec:method-anchors}
\paragraph{Anchor placement.}
Let \(\mathcal{T}=[t_1,\ldots,t_L]\) be a sequence of \(L\) text tokens.
An anchoring policy specifies an ordered set of text boundaries
\(\mathcal{B}=\{b_m\}_{m=1}^{M}\), where
\(0=b_0<b_1<\cdots<b_M\leq L\). We insert an anchor position after each
boundary:
\begin{equation}
  \widehat{\mathcal{T}}
  =
  \mathop{\mathbin{\Vert}}_{m=1}^{M}
  \Big(
    [t_{b_{m-1}+1},\ldots,t_{b_m}]
    \mathbin{\Vert}[\xi_m]
  \Big)
  \mathbin{\Vert}
  [t_{b_M+1},\ldots,t_L].
  \label{eq:march-anchors}
\end{equation}
where \(\Vert\) denotes sequence concatenation, and \(\xi_m\) is the
\(m\)-th occurrence of a shared learned anchor embedding \(\xi\). Text and anchor positions serve different computational roles. Text
positions apply the base recurrent update, allowing the matrix-valued state
to evolve continuously across anchor boundaries. Immediately after
processing \(t_{b_m}\), MARCH checkpoints the resulting cumulative state
to form the \(m\)-th state anchor. The following anchor position \(\xi_m\)
does not modify the recurrent state; instead, its hidden representation
provides the routing metadata associated with that checkpoint. Thus, each
anchor boundary produces two coupled objects: a snapshot of the recurrent
memory and a compact representation through which that snapshot can later
be retrieved. We formalize these two operations next.

\paragraph{Cumulative recurrent-state checkpointing.}
At layer \(\ell\), MARCH leaves the underlying recurrent update unchanged
and carries the state
\(\mathbf{S}^{(\ell)}_t\in\mathbb{R}^{d_v\times d_k}\)
continuously across anchor boundaries. At each boundary \(b_m\), it
snapshots the current state:
\begin{equation}
  \mathbf{A}^{(m,\ell)}
  =
  \mathbf{S}^{(\ell)}_{b_m}
  \in\mathbb{R}^{d_v\times d_k},
  \qquad m=1,\ldots,M.
  \label{eq:march-anchor}
\end{equation}
Because the recurrence is not reset between anchor boundaries,
\(\mathbf{A}^{(m,\ell)}\) encodes the cumulative prefix up to position
\(b_m\), rather than only the segment since the preceding boundary. We
therefore refer to it as a \emph{state anchor}. The ordered bank
\(\{\mathbf{A}^{(1,\ell)},\ldots,\mathbf{A}^{(M,\ell)}\}\) traces the
temporal evolution of a single recurrent memory, preserving earlier
versions before subsequent decay and delta updates attenuate or modify
their contents.

\paragraph{Content-conditioned anchor metadata.}
Let $\mathbf{u}^{(\ell)}_m$ denote the normalized input representation of
anchor position $\xi_m$ at layer $\ell$. The anchor position reads only
its aligned state checkpoint:
\begin{equation}
  \mathbf{q}^{(\ell)}_m
  =
  \mathbf{W}^{(\ell)}_q\mathbf{u}^{(\ell)}_m,
  \qquad
  \mathbf{o}^{(\ell)}_m
  =
  \mathbf{A}^{(m,\ell)}\mathbf{q}^{(\ell)}_m.
  \label{eq:march-anchor-token-read}
\end{equation}
The same input representation is projected into a compact routing key:
\begin{equation}
  \boldsymbol{\kappa}^{(\ell)}_m
  =
  \mathbf{W}^{(\ell)}_k
  \mathbf{u}^{(\ell)}_m
  \in\mathbb{R}^{d_r}.
  \label{eq:march-routing-key}
\end{equation}
The aligned readout is incorporated into the anchor position through
the standard output projection and residual pathway. Consequently,
$\mathbf{u}^{(\ell+1)}_m$ depends on
$\mathbf{A}^{(m,\ell)}$, and the routing key produced at layer
$\ell+1$ becomes conditioned on the content retained by the aligned
state anchor. Thus, although all anchor positions share the same
learned input embedding, they acquire distinct, state-dependent
representations after the first layer. This cross-layer construction makes routing explicitly dependent on
what each state anchor contains, rather than only on its temporal
index.

\subsection{Content-Routed Historical Reading}
\label{sec:method-routing}
\label{sec:method-history-read}

\paragraph{Content-based routing.}
For a text token at position $t$, the causally available state anchors are
indexed by $\mathcal{V}_t=\{m\in\{1,\ldots,M\}\mid b_m<t\}$. MARCH projects
the normalized hidden state $\mathbf{x}_t$ into a routing query and scores it
against the key of each visible anchor:
\begin{equation}
  \boldsymbol{\rho}_t
  =
  \mathbf{W}_R\mathbf{x}_t,
  \qquad
  a_{t,m}
  =
  \boldsymbol{\rho}_t^{\top}\boldsymbol{\kappa}_m,
  \quad m\in\mathcal{V}_t.
  \label{eq:march-routing-score}
\end{equation}
To allow the model to bypass historical memory, we augment the visible
anchor set with a null option $\varnothing$, whose payload is fixed to
zero, $\mathbf{A}^{(\varnothing)}=\mathbf{0}$. Its query-dependent
logit is $n_t=\mathbf{w}_{\varnothing}^{\top}\mathbf{x}_t
+b_{\varnothing}$.
Let $\widetilde{\mathcal{V}}_t
=\mathcal{V}_t\cup\{\varnothing\}$ denote the augmented candidate set.
We define the logit of each candidate $j\in\widetilde{\mathcal{V}}_t$ as
\begin{equation}
  s_{t,j}
  =
  \begin{cases}
    a_{t,j}, & j\in\mathcal{V}_t,\\
    n_t,     & j=\varnothing,
  \end{cases}
  \qquad
  \pi_{t,j}
  =
  \frac{\exp(s_{t,j})}
       {\displaystyle
        \sum_{r\in\widetilde{\mathcal{V}}_t}
        \exp(s_{t,r})}.
  \label{eq:march-routing-prob}
\end{equation}
Since the selected routing probabilities directly weight the historical state readouts, their scores remain jointly optimized by the language-modeling objective. The routing query $\boldsymbol{\rho}_t$ determines which anchors to retrieve, whereas the state-read query $\mathbf{q}_t$ reads
their matrix-valued contents. If no anchor is visible, the null option
receives all probability mass.

We note that the aggregation formulation in Equation \eqref{eq:march-routing-prob} readily admits a sparse variant by restricting aggregation to the $K$ highest-scoring visible anchors (Top-$K$). This sparse approach exhibits natural connections with the hierarchical sparse attention approaches \citep{lu2025moba, HiLS}, while preserving dense token-level processing rather than relying on hard token-level pruning. Our ablation studies in Section \ref{sec:ablations} show that sparse routing substantially reduces aggregation cost with minimal performance degradation.

\paragraph{Historical retrieval and residual fusion.}
Given the routing probabilities, the causally visible state anchors
are aggregated into a query-dependent historical state, which is read
using the same state-read query as the current state. The resulting historical
readout is then added to the current-state readout:
\begin{equation}
  \mathbf{o}_t
  =
  \mathbf{S}_t\mathbf{q}_t
  +
  \sum_{j\in\widetilde{\mathcal{V}}_t}
  \pi_{t,j}
  \mathbf{A}^{(j)}\mathbf{q}_t.
  \label{eq:march-core}
\end{equation}
This additive formulation preserves the original recurrent path and
introduces historical retrieval as an auxiliary residual branch,
without modifying the underlying recurrent update. Since the routing probabilities
directly affect the layer output, the routing queries and anchor-derived
keys are optimized end-to-end with the language-modeling objective.

\subsection{Implementation}
\label{sec:method-efficient-training}
We implement MARCH as a two-stage producer--reader computation. Following the hardware-efficient chunkwise formulation of Gated DeltaNet \citep{yang2025gateddelta}, the producer processes recurrent updates in blocks amenable to tensor-core acceleration, computes each token's current-state output, and checkpoints the recurrent state at each anchor boundary. The resulting state anchors are consumed by the historical reader. Inspired by the I/O-aware principles of FlashAttention \citep{dao2022flashattention}, the reader jointly tiles query tokens and state anchors, reuses each anchor tile across a block of queries, and fuses routing-score computation, online softmax updates, and the accumulation of weighted state readouts into a streaming reduction. This fused schedule avoids materializing either the dense token-to-anchor routing matrix or the substantially larger tensor of per-anchor candidate readouts, thereby reducing intermediate storage and the associated HBM traffic. As shown in \Cref{fig:training-efficiency}, despite the cost of historical retrieval, our fused dense implementation exceeds FlashAttention-2 in throughput at 64K and above and incurs lower core runtime from 32K onward.

\section{Experiments}
\label{sec:experiments}

\sysname{} is designed to extend the long-range memory of recurrent models
while preserving their general language capabilities.  
In this paper, we verify the effectiveness of \sysname{} by training from scratch, and evaluate across a diverse suite of benchmarks  spanning zero-shot commonsense reasoning,
long-context understanding, and in-context retrieval. Across these tasks,
\sysname{} consistently outperforms existing recurrent baselines, with
particularly strong gains on retrieval-intensive and long-context benchmarks.

\subsection{Experimental Setup}
\label{sec:experimental-setup}

\paragraph{Training configuration.}
Following the academic-scale protocol used by Log-Linear Attention~\citep{guo2026loglinear}, we pretrain
the models from scratch on 50B tokens from the
Long-Data-Collections dataset, using a sequence length of 16K.  The main configurations use 21 layers and a
hidden size of 1536.  The
Transformer (693M) uses 16 attention heads and a RoPE base of 500K, while Gated DeltaNet 
(793M) and its variants use six value heads. To control for parameter count in
addition to model depth, we also include a \(24\)-layer Transformer with
(778M) parameters, closely matching the size of the Gated DeltaNet. For
\sysname{}, we set the routing dimension to \(d_r=64\) and use a
periodic anchoring interval of \(C=512\) text tokens.
We train all models with a global batch size of approximately \(4.2\)M tokens
using the fused AdamW optimizer, with \(\beta_1=0.9\), \(\beta_2=0.95\),
\(\epsilon=10^{-8}\), and a weight decay of \(0.1\). The peak learning rate is
set to \(4\times10^{-4}\) with a warmup-stable-decay schedule.
All models use the same
training data, token budget, context length, and optimization configuration.
\paragraph{Baselines.}
Our primary comparisons are against standard GDN ~\citep{yang2025gateddelta} and GDN augmented with
Log-Linear Attention~\citep{guo2026loglinear}. \sysname{} and these two
baselines use matched architectural configurations and the same pretraining
setup, enabling a controlled comparison of their memory mechanisms. To
contextualize their performance against full attention, we additionally
include two Transformer baselines: a \(21\)-layer model matched in depth to
the recurrent models and a \(24\)-layer model approximately matched to them
in parameter count.

\paragraph{Evaluation tasks.}
For short-context generalization, we use eight zero-shot commonsense
benchmarks: LAMBADA~\citep{paperno_lambada_2016},
PIQA~\citep{bisk_piqa_2020}, HellaSwag~\citep{zellers_hellaswag_2019},
WinoGrande~\citep{sakaguchi_winogrande_2021}, ARC-Easy and
ARC-Challenge~\citep{clark_think_2018}, OpenBookQA~\citep{OpenBookQA2018},
and CommonsenseQA~\citep{talmor2019commonsenseqaquestionansweringchallenge}.
We additionally evaluate long-context understanding on
LongBench~\citep{bai2024longbench}, covering single-document QA,
multi-document QA, summarization, and few-shot learning.  The long-context
retrieval evaluation covers six single-neddle and multi-needle tasks from
RULER~\citep{hsieh2024ruler} at 4K, 8K, and 16K context lengths.  Finally, the
in-context retrieval suite contains SQuAD~\citep{rajpurkar_know_2018},
TriviaQA~\citep{JoshiTriviaQA2017}, SWDE~\citep{lockard_openceres_2019},
FDA~\citep{arora_language_2023}, Natural
Questions~\citep{kwiatkowski-etal-2019-natural}, and
DROP~\citep{dua2019drop}.  We follow the evaluation protocol of prior work~\citep{wang2026dynamic} and
use the LM-Evaluation-Harness~\citep{eval-harness}. 

\begin{table*}[t]
  \centering
  \caption{Zero-shot performance of MARCH and baseline models on eight
commonsense reasoning benchmarks. Results are reported using accuracy
(\texttt{acc}) or normalized accuracy (\texttt{acc\_n}), as indicated in
the column headers; higher is better
(\textcolor{ForestGreen}{$\uparrow$}). The best result among Gated DeltaNet variants in each column is highlighted in bold.}
  \label{tab:lm_cs}
  \small
  \begin{tabular}{r|cccccccc|c}
      \toprule
      \textbf{Model}
      & \textbf{LMB.} & \textbf{PIQA} & \textbf{Hella.} & \textbf{Wino.}
      & \textbf{ARC-e} & \textbf{ARC-c} & \textbf{OBQA} & \textbf{CSQA}
      & \textbf{Avg.} \\
      & acc \textcolor{ForestGreen}{$\uparrow$}
      & acc \textcolor{ForestGreen}{$\uparrow$}
      & acc \textcolor{ForestGreen}{$\uparrow$}
      & acc \textcolor{ForestGreen}{$\uparrow$}
      & acc\_n \textcolor{ForestGreen}{$\uparrow$}
      & acc\_n \textcolor{ForestGreen}{$\uparrow$}
      & acc\_n \textcolor{ForestGreen}{$\uparrow$}
      & acc \textcolor{ForestGreen}{$\uparrow$}
      & \\
      \midrule
      Transformer
      & 49.4 & 66.5 & 33.9 & 52.1 & 47.8
      & 26.4 & 32.0 & 22.1 & 41.3 \\
      w/ \emph{24 Layers}
      & 50.3 & 67.6 & 34.4 & 50.6 & 46.3
      & 25.8 & 31.2 & 24.7 & 41.4 \\
      \midrule
      Gated DeltaNet
      & 48.5 & 66.1 & 33.1 & 50.8 & 45.9 & 25.3
      & 30.0 & 21.1 & 40.1 \\
      w/ \emph{Log-Linear}
      & 47.7 & 65.7 & 33.2 & 51.9 & 44.3 & 24.9
      & 30.4 & 21.7 & 40.0 \\
      w/ \emph{\sysname}
      & \textbf{49.5} & \textbf{66.9} & \textbf{34.8} & \textbf{52.6}
      & \textbf{47.1} & \textbf{25.6} & \textbf{32.8} & \textbf{22.5}
      & \textbf{41.5} \\
      \bottomrule
  \end{tabular}
\end{table*}

\subsection{Main Results}
\label{sec:experimental-results}

\paragraph{Commonsense reasoning.}
As shown in \Cref{tab:lm_cs}, \sysname{} consistently outperforms both the
vanilla and Log-Linear variants of Gated DeltaNet across all eight zero-shot
commonsense reasoning benchmarks. It improves the average accuracy from
$40.1$ and $40.0$ to $41.5$, respectively, with the largest gain over the
vanilla baseline observed on OpenBookQA ($+2.8$ points), aligning with our findings in retrieval tasks presented below. Moreover, \sysname{}
achieves a higher average score than both Transformer baselines, surpassing
the standard Transformer on six of eight tasks and the 24-layer Transformer
on four. These results indicate that \sysname{} consistently strengthens the
Gated DeltaNet backbone while remaining competitive with comparable
full-attention models on short-context language understanding tasks.

\begin{figure}[!hb]
  \centering
  \includegraphics[width=\textwidth]{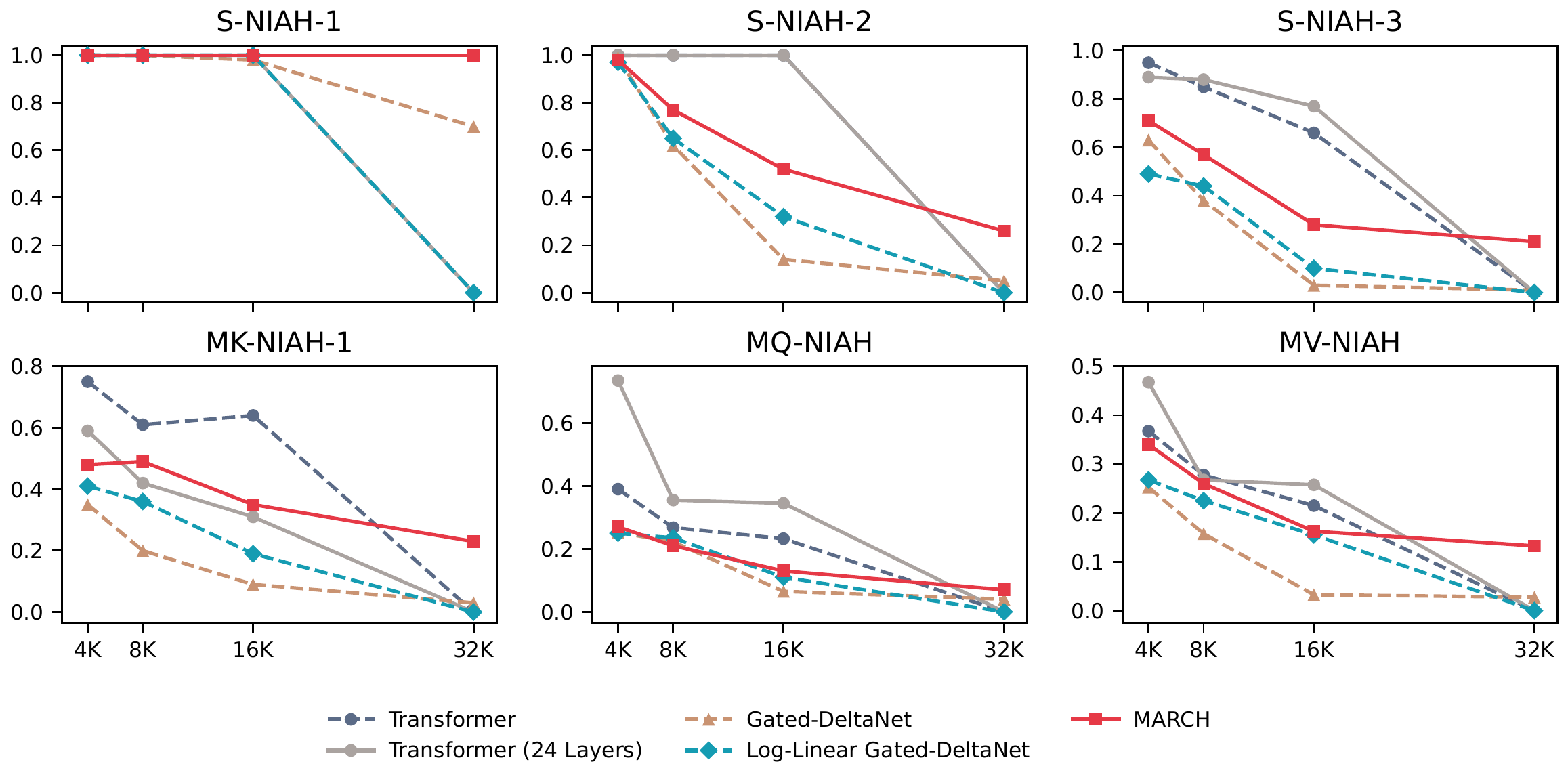}
  \caption{NIAH performance on three single-needle and three multi-needle tasks. The Transformer achieves perfect accuracy on both S-NIAH-1 and S-NIAH-2 at context lengths of 4K, 8K, and 16K.}
  \label{fig:niah-lineplot}
\end{figure}

\paragraph{Needle-in-a-haystack retrieval.}
We evaluate long-context associative retrieval using the needle-in-a-haystack
(NIAH) suite from RULER~\citep{hsieh2024ruler}, where a model must recover
values associated with keys embedded among irrelevant context. All models are trained with a maximum context
length of 16K; \cref{fig:niah-lineplot} reports results from 4K to 32K, making 32K a zero-shot length-extrapolation setting. Across the 24 task--length combinations, \sysname{} outperforms the
stronger recurrent baseline in 19 settings and matches it in the remaining
five. On the multi-needle tasks, it wins in 11
of 12 settings. At 32K, \sysname{} achieves the best result on all six
tasks, retaining perfect accuracy on S-NIAH-1 and nonzero accuracy on
the remaining tasks, whereas both Transformer variants and Log-Linear
Gated-DeltaNet score zero throughout. This contrast is consistent with
RoPE extrapolation in the Transformers and the state-index-dependent
coefficients of Log-Linear Gated-DeltaNet. MARCH instead shares the same
content-based router across all anchors, allowing longer contexts to
introduce additional anchors without requiring new anchor-specific
routing parameters.

\begin{table*}[t]
\centering
\caption{Results on twelve LongBench tasks. The
best result among Gated DeltaNet and its variants is shown in bold for each
task. The relative average gain over the Gated DeltaNet is shown in
green parentheses.}
\label{tab:long_bench}
\small
\setlength{\tabcolsep}{3.2pt}
\renewcommand{\arraystretch}{1.08}
\begin{tabular}{@{}l|ccc|ccc|ccc|ccc|c@{}}
\toprule
& \multicolumn{3}{c|}{\textbf{Single-Doc QA}} & \multicolumn{3}{c|}{\textbf{Multi-Doc QA}} & \multicolumn{3}{c|}{\textbf{Summarization}} & \multicolumn{3}{c|}{\textbf{Few-shot Learning}} & \\ \midrule
 \textbf{Model} & {NQA} & {QQA} & {MFQ} 
          & {HQA} & {2WM} & {Mus} 
          & {GvR} & {QMS} & {MNs}
          & {TRC} & {TQA} & {SSM}
          & \textbf{Avg.\textcolor{ForestGreen}{$\uparrow$}} \\\midrule
Transformer               & 4.4 & 4.1 & 15.9 & 7.5 & 9.9 & 4.1 & 10.7 & 11.6 & 14.5 & 21.0 & 33.9 & 28.3 & 13.8\\
\quad w/ \emph{24 Layers} & 3.5 & 11.1 & 18.3 & 7.8 & 9.8 & 4.1 & 11.3 & 12.9 & 12.8 & 22.5 & 47.5 & 23.2 & 15.4\\
\midrule
Gated DeltaNet             & 3.0 & 4.8 & 13.3 & 5.6 & 8.7 & 2.1 & 2.6 & 11.3 & 13.0 & 18.0 & 38.6 & 21.9 & 11.9\\
\quad w/ \emph{Log-Linear} & 3.6 & 6.2 & 13.6 & 7.1 & 8.2 & 3.3 & 6.3 & 13.2 & 13.5 & 17.0 & 32.8 & 25.1 & 12.5\\
\quad w/ \sysname          &
\textbf{4.2} &
\textbf{7.8} &
\textbf{14.6} &
\textbf{7.4} &
\textbf{11.5} &
\textbf{4.8} &
\textbf{8.2} &
\textbf{17.4} &
\textbf{14.1} &
\textbf{19.0} &
\textbf{43.1} &
\textbf{26.3} &
\textbf{14.9 (\textcolor{ForestGreen}{$\uparrow25\%$})}\\
\bottomrule
\end{tabular}
\end{table*}

\paragraph{Long-context understanding.}
\Cref{tab:long_bench} reports results across four LongBench task categories.
\sysname{} consistently outperforms both vanilla Gated DeltaNet  and its log-linear variant
on all twelve tasks. The improvements are particularly pronounced on
multi-document QA: relative to the stronger of the vanilla and log-linear
Gated DeltaNet baselines, \sysname{}
raises the 2WikiMultihopQA score from 8.7 to 11.5 and the MuSiQue score from 3.3
to 4.8, corresponding to relative gains of 32\% and 45\%, respectively. The
benefits also extend to summarization, where the QMSum score increases from
13.2 to 17.4 (32\%), and to all three few-shot learning tasks. These results
show that content-routed state anchors improve long-context understanding
across diverse task formats, rather than benefiting only retrieval-oriented
question answering.

\begin{table*}[t]
\centering
\caption{In-context retrieval accuracy
(\textcolor{ForestGreen}{$\uparrow$}). The best result among Gated DeltaNet and
its variants is marked in bold for each benchmark, with the relative gain over
the stronger of the two Gated DeltaNet baselines shown in green parentheses.}
\label{tab:in-context-retrieval}
\resizebox{\linewidth}{!}{%
\begin{tabular}{r|cccccc|c}
\toprule
\textbf{Model}
& \textbf{SQuAD\textcolor{ForestGreen}{$\uparrow$}} 
& \textbf{SWDE\textcolor{ForestGreen}{$\uparrow$}} 
& \textbf{FDA\textcolor{ForestGreen}{$\uparrow$}} 
& \textbf{TriviaQA\textcolor{ForestGreen}{$\uparrow$}}
& \textbf{DROP\textcolor{ForestGreen}{$\uparrow$}}
& \textbf{NQ\textcolor{ForestGreen}{$\uparrow$}} 
& \textbf{Avg.\textcolor{ForestGreen}{$\uparrow$}} \\
\midrule
Transformer & 41.3 & 59.3 & 80.4 & 2.2 & 2.9 & 1.7 & 31.3 \\
w/ \emph{24 Layers} & 40.4 & 64.9 & 83.7 & 4.0 & 3.4 & 2.5 & 33.2 \\
\midrule
Gated DeltaNet & 34.8 & 45.0 & 31.4 & 1.1 & 2.2 & 0.8 & 19.2 \\
w/ \emph{Log-Linear} & 33.7 & 46.1 & 38.2 & 1.3 & 2.6 & 1.0 & 20.5 \\
w/ \emph{\sysname}
& \textbf{37.7 (\textcolor{ForestGreen}{$\uparrow8\%$})}
& \textbf{51.9 (\textcolor{ForestGreen}{$\uparrow13\%$})}
& \textbf{44.6 (\textcolor{ForestGreen}{$\uparrow17\%$})}
& \textbf{1.6 (\textcolor{ForestGreen}{$\uparrow23\%$})}
& \textbf{2.9 (\textcolor{ForestGreen}{$\uparrow12\%$})}
& \textbf{1.2 (\textcolor{ForestGreen}{$\uparrow20\%$})}
& \textbf{23.3 (\textcolor{ForestGreen}{$\uparrow14\%$})} \\
\bottomrule
\end{tabular}%
}
\end{table*}

\paragraph{In-Context Retrieval.}
Following~\citep{arora2024simple}, we evaluate in-context retrieval on six
real-world, recall-intensive benchmarks. As shown in
\cref{tab:in-context-retrieval}, \sysname{} consistently outperforms both
vanilla Gated DeltaNet and its Log-Linear variant across all tasks. Relative
to the stronger of the vanilla and log-linear Gated DeltaNet baselines on each
benchmark, \sysname{} yields
relative improvements ranging from $8\%$ on SQuAD to $23\%$ on TriviaQA and
raises the average accuracy from $20.5$ to $23.3$, corresponding to a
$14\%$ relative improvement. These consistent gains across heterogeneous
retrieval tasks demonstrate that \sysname{} improves the retrieval capability
of the Gated DeltaNet backbone beyond a particular dataset or input format.
Together, these results establish content-routed state anchors as an effective
mechanism for strengthening fine-grained retrieval in recurrent models.

\begin{center}
  \begin{minipage}{\textwidth}
    \centering
    \includegraphics[width=\linewidth]{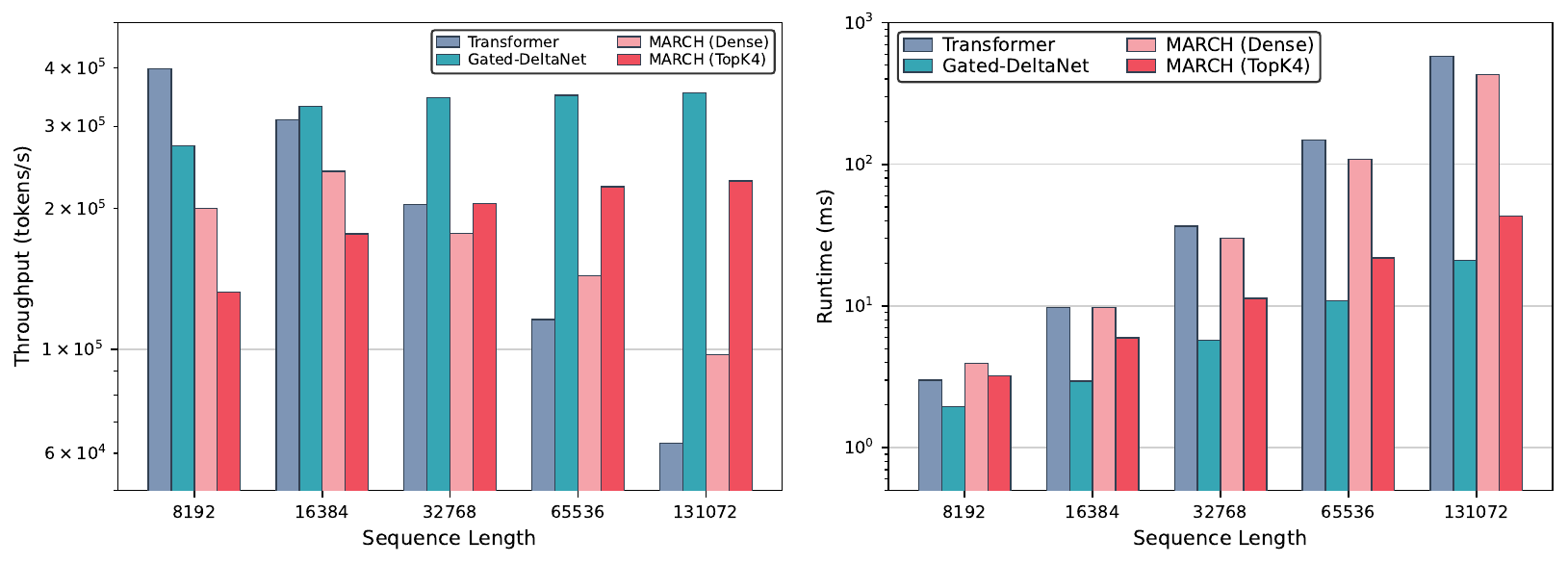}
    \captionsetup{hypcap=false}
    \captionof{figure}{Training efficiency across sequence lengths. Left:
    end-to-end training throughput in tokens per second (higher is better).
    Right: forward--backward runtime of the core sequence-mixing operation in
    milliseconds (lower is better). \sysname{} (Top-$4$) retains only the four
    highest-scoring state anchors for each token and head during historical
    retrieval.}
    \label{fig:training-efficiency}
  \end{minipage}
\end{center}

\paragraph{Training efficiency.}
\Cref{fig:training-efficiency} compares the end-to-end throughput and core
forward--backward runtime of FlashAttention-2, Gated DeltaNet, dense
\sysname{}, and its Top-$4$ implementation. Sparse routing becomes
increasingly beneficial as the context grows. At 128K tokens, Top-$4$
\sysname{} more than doubles the training throughput of dense \sysname{} and
reduces its core runtime by roughly an order of magnitude. It also achieves
higher throughput than FlashAttention-2 at this length, although vanilla
Gated DeltaNet remains faster because it incurs no historical-retrieval
overhead.

\section{Ablation Studies}
\label{sec:ablations}

\paragraph{Effect of chunk size.}
The training chunk size $C$ determines how frequently \sysname{} checkpoints
the recurrent state. Smaller chunks create denser candidate anchors and offer
finer temporal resolution, at the cost of a larger anchor cache and higher
historical-routing overhead. We vary $C$ from 256 to 2048 while keeping all
other model and training settings fixed. \Cref{tab:chunk-size-ablation} reports performance on three in-context
retrieval benchmarks and the average over the six NIAH tasks at context
lengths of 4K, 8K, and 16K. As an additional inference test, we organize the \sysname{} state bank
according to the Fenwick tree scheme used by Log-Linear
Attention~\citep{guo2026loglinear,Fenwick1994AND}. This scheme arranges state
anchors hierarchically and retains $\mathcal{O}(\log T)$ anchors as the context
grows. 
We report performance close to \citep{guo2026loglinear}. This demonstrates that the learned router in \sysname{} shows great generalizability and flexibility across various state bank organization schemes.

\begin{center}
  \begin{minipage}{\linewidth}
  \centering
  \captionof{table}{Effect of chunk size on in-context retrieval and NIAH.
  Panel (a) evaluates checkpoints using the same chunk size during training
  and inference. Panel (b) varies the inference-time chunk size for the
  checkpoint trained with chunk size 512. NIAH scores are averaged over six tasks at each context
  length. Bold indicates the selected setting in each panel and the best
  result in each column.}
  \label{tab:chunk-size-ablation}
  \small
  \setlength{\tabcolsep}{7pt}
  \begin{tabular}{cc|ccc|ccc}
    \toprule
    & & \multicolumn{3}{c|}{\textbf{In-context Retrieval}}
    & \multicolumn{3}{c}{\textbf{NIAH}} \\
    \textbf{Chunk Size}
    & \textbf{\# Anchors}
    & \textbf{SQuAD}
    & \textbf{SWDE}
    & \textbf{FDA}
    & \textbf{4K}
    & \textbf{8K}
    & \textbf{16K} \\
    \midrule
        \multicolumn{8}{l}{\textbf{(a) Matched training and inference chunk sizes}} \\
        \addlinespace[1pt]
    256
    & 64 & 36.76 & 49.23 & \textbf{44.83}
    & 58.17 & 49.25 & \textbf{44.83} \\
    \textbf{512}
    & 32 & \textbf{37.70} & \textbf{51.85} & 44.56
    & \textbf{59.58} & \textbf{54.96} & 39.46 \\
    1024
    & 16 & 36.49 & 48.51 & 43.28
    & 53.21 & 43.21 & 33.54 \\
    2048
    & 8 & 34.15 & 44.19 & 30.76
    & 59.54 & 39.83 & 27.96 \\
    \midrule
    \multicolumn{8}{l}{\textbf{(b) Inference-time chunk size }} \\
    \addlinespace[1pt]
    64
    & 256 & 39.41 & 53.11 & 46.01
    & \textbf{63.33} & 55.00 & 39.67 \\
    128
    & 128 & 39.08 & 52.84 & 46.91
    & \textbf{63.33} & \textbf{57.33} & 39.17 \\
    \textbf{256}
    & 64 & \textbf{40.35} & \textbf{53.38} & \textbf{47.46}
    & 62.83 & 56.17 & \textbf{40.83} \\
    512
    & 32 & 37.70 & 51.85 & 44.56
    & 59.58 & 54.96 & 39.46 \\
    1024
    & 16 & 37.23 & 43.74 & 34.57
    & 55.33 & 48.33 & 32.67 \\
    2048
    & 8 & 37.23 & 33.93 & 23.23
    & 51.00 & 40.33 & 34.00 \\
    Fenwick Tree
    & 6 & 36.16 & 36.79 & 22.35
    & 52.17 & 39.56 & 33.14 \\
    \bottomrule
  \end{tabular}
  \end{minipage}
\end{center}

In Panel (a), $C=512$ provides the best overall balance between retrieval
quality and anchor count. Smaller chunks improve some long-context results
but incur higher memory and routing costs, whereas larger chunks generally
degrade retrieval because the resulting checkpoints are too sparse. We
therefore adopt $C=512$ as the default. Panel (b) shows that changing the
chunk size at inference provides a flexible accuracy--memory trade-off:
denser anchors generally improve retrieval at higher cost, while overly
sparse anchors lead to substantial degradation. The Fenwick tree row
additionally evaluates hierarchical organization of the state bank at
inference.
\paragraph{Routing design.}
We ablate the router's query--key dimension $d_r$, routing sparsity, and
learned null option. \Cref{tab:routing-ablation} reports aggregate results
across general language understanding, long-context benchmarks, and NIAH. The
default uses dense routing with $d_r=64$ and includes the null option.

\begin{center}
  \begin{minipage}{\linewidth}
  \centering
  \captionof{table}{Routing-design ablations. The first row is the default;
  each subsequent row changes one component. Commonsense (CS), LongBench,
  and Retrieval are macro-averages over 8, 12, and 6 benchmarks,
  respectively. NIAH scores are averaged over six tasks, and Avg. over the
  three context lengths. Column-wise best results are bold.}
  \label{tab:routing-ablation}
  \small
  \setlength{\tabcolsep}{3.5pt}
  \begin{tabular}{@{}lcc|ccc|cccc@{}}
    \toprule
    \multicolumn{3}{c|}{\textbf{Configuration}}
    & \multicolumn{3}{c|}{\textbf{Benchmark Averages}}
    & \multicolumn{4}{c}{\textbf{NIAH}} \\
    \textbf{Routing}
    & \textbf{$d_r$}
    & \textbf{Null}
    & \textbf{CS}
    & \textbf{LongBench}
    & \textbf{Retrieval}
    & \textbf{4K}
    & \textbf{8K}
    & \textbf{16K}
    & \textbf{Avg.} \\
    \midrule
    Dense (default) & 64 & Yes
    & \textbf{41.48} & \textbf{14.87} & 23.31
    & \textbf{59.58} & \textbf{54.96} & \textbf{39.46} & \textbf{51.33} \\
    \midrule
    Dense & 192 & Yes
    & 40.94 & 13.88 & \textbf{24.52}
    & 56.71 & 51.33 & 37.38 & 48.47 \\
    Top-$4$ & 64 & Yes
    & 41.38 & 13.79 & 23.17
    & 57.29 & 46.04 & 31.21 & 44.85 \\
    Dense & 64 & No
    & 41.04 & 14.11 & 22.86
    & 54.88 & 47.92 & 35.13 & 45.98 \\
    \bottomrule
  \end{tabular}
  \end{minipage}
\end{center}

Increasing $d_r$ to 192 yields higher retrieval capability but reduces general performance across other tasks, making $d_r=64$ a more balanced choice overall. Top-$4$ nearly matches dense routing
on commonsense and retrieval but trails on NIAH, making it an
efficiency-oriented operating point when considered alongside
\Cref{fig:training-efficiency}. Removing the null option degrades every
aggregate, confirming the benefit of bypassing irrelevant historical states. 

\section{Related Work}
\label{sec:related-work}

\paragraph{Efficient Attention Mechanisms.}
Efficient attention reduces the quadratic cost of full self-attention through
local windows, kernelization, or systems optimization.  Local sliding-window
attention limits each query to a bounded neighborhood
\citep{wang2025rattentionminimalslidingwindow,
cabannes2025shortwindowattentionenables}.  Performer, Nystromformer, and Linear
Attention replace the softmax kernel with feature maps and exploit associativity
for linear-time computation
\citep{performer,xiong2021nystromformer,katharopoulos2020transformers}.
FlashAttention-2, sequence parallelism, and chunkwise algorithms instead
improve hardware efficiency without changing the dense attention pattern
\citep{flashattention2,Sun2024LinearAS}.

\paragraph{Sparse Attention.}
Sparse attention retains content-based softmax retrieval but restricts each
query to a small subset of token-level key--value pairs.  Early methods rely on
predefined connectivity: Sparse Transformer factorizes the attention pattern,
while Longformer and BigBird combine local windows with global or random links
\citep{child2019generating,beltagy2020longformer,zaheer2020big}.  Later methods
make the sparse pattern input dependent: Routing Transformer clusters tokens by
content, H$_2$O evicts low-utility cache entries, and Quest selects KV-cache
pages conditioned on the current query
\citep{roy2021efficient,zhang2023h2o,tang2024quest}.  More recent trainable
designs route queries to relevant blocks, as in MoBA, or combine compressed,
selectively retrieved, and local branches with hardware-aligned kernels, as in
Native Sparse Attention~\citep{lu2025moba,yuan2025native}.  These approaches
reduce attention-score computation or memory traffic, but their accuracy hinges
on token or block selection and they still store or manipulate token-level KV
memories.  In contrast, MARCH routes over compact keys associated with
historical recurrent-state snapshots, retrieving compressed prefix states
rather than sparsifying token-to-token attention.

\paragraph{State Space Models and Gated Linear Recurrences.}
State space models (SSMs) and linear recurrent networks compress the prefix into
a recurrent state.  Linear Attention and its kernelized variants share this
view through decayed outer-product updates and query-based reads
\citep{katharopoulos2020transformers,chou2024metala,dao2024transformers}.
S4 uses structured linear dynamics, while Mamba and Mamba-2 use selective
transitions; RetNet, RWKV, HGRN, and LRU combine associative memories with
structured recurrences
\citep{gu2021efficiently,gu2023mamba,dao2024transformers,sun2023retentive,peng2023rwkv,
qin2024hgrn2,orvieto2023resurrecting,longhorn}.  GLA introduces input-dependent
decay, DeltaNet and GDN use delta-rule corrections, and GDN-2 decouples erase
and write through channel-wise gates
\citep{yang2025gateddelta,yang2024parallelizing,
hatamizadeh2026gateddeltanet2,siems2025deltaproduct,Grazzi2024UnlockingSI}.
Despite these advances, most models retain one fixed-capacity state, whose
dimension directly controls update and read cost; this bottleneck contributes
to the retrieval gap with Transformers
\citep{arora2024simple,wen_rnns_2024}.  We preserve efficient recurrence while
expanding memory into selectively accessed states.

\paragraph{State Expansion and Associative Memory.}
Long-context studies identify recurrent state capacity as a central limitation
of Linear Attention models~\citep{arora2024zoology,arora2024simple}.
Multi-State RNNs, HGRN2, and Log-Linear Attention expand or hierarchically
organize recurrent states~\citep{Oren2024TransformersAM,qin2024hgrn2,
guo2026loglinear}.  Mixture-of-Memories, Sparse State Expansion, Product Key
Memory, and Fast-weight Product Key Memory use memory experts or sparse banks,
while Sparse Delta Memory sparsifies GDN reads and writes
\citep{du2025mom,pan25SSE,lample2019largememorylayersproduct,
berges2024memorylayersscale,zhao2026fastweightproductkeymemory,
cabannes26SDM,afzal2026raven}.  Context-compression methods instead retrieve
at the token or chunk level, using learned summary tokens or selective chunk
reopening~\citep{chevalier2023adapting,mu2023learning,zhang2024long,
deng2025unigist,petrov2025long,mao2026gisttokens}.  Our method treats
recurrent states as retrieval units, expands total capacity, and reads only selected states, decoupling capacity from dense per-token updates.

\section{Limitations and Future Work}
\label{sec:limitations-and-future-work}
MARCH adopts periodic checkpointing and organizes all historical
states in a single homogeneous anchor bank. Although this design is simple and
efficient, fixed-interval anchoring does not account for the non-uniform
evolution of recurrent memory: it may create redundant anchors in stable
regions while providing insufficient resolution when the state changes
rapidly. A natural extension is to develop adaptive anchoring mechanisms according to state novelty
or update magnitude and to consolidate or evict redundant anchors given a memory budget. More broadly, MARCH improves access to earlier states but
does not explicitly increase or specialize the capacity of the underlying
memory. Future work could combine state anchoring with larger-capacity memory
and multiple memory partitions specialized for different temporal scales or
information types. For example, short-term context, salient episodic events,
and slowly consolidated knowledge could be maintained through distinct write,
retention, and forgetting mechanisms, while a hierarchical router
determines both which memory partition and which stored state should serve
each query.
In addition, MARCH has the potential to support external memory modules for optimized performance over specific downstream tasks, knowledge consolidation from experience to parametric information, and other memory manipulation mechanisms, opening up new scaling directions for test-time training and continual learning.
\section{Conclusion}
\label{sec:conclusion}

We introduce \sysname{}, a novel attention architecture which augments recurrent models with content-routed state anchors. By preserving cumulative state checkpoints, \sysname{} enables
selective access to earlier recurrent states without modifying the underlying
recurrence. It consistently outperforms strong recurrent baselines across
commonsense reasoning, LongBench, in-context retrieval, and NIAH. Ablations further
show a controllable retrieval--efficiency trade-off through checkpoint density
and sparse routing. These results establish historical-state retrieval as a
practical approach to scaling recurrent memory beyond a single evolving state.

\bibliography{main}


\end{document}